\documentclass[conference,a4paper]{IEEEtran}
\usepackage[T1]{fontenc}
\usepackage[utf8]{inputenc}
\usepackage{amsmath,amssymb}
\usepackage{graphicx}
\usepackage{cite}
\usepackage{booktabs}
\usepackage{url}
\usepackage{xcolor}
\usepackage{tikz}
\usepackage{hyperref}
\newcommand{\dmin}{d_{\min}}
\usetikzlibrary{arrows.meta,positioning,fit,backgrounds,calc}
\title{LACART: Lookahead-Augmented Collision Avoidance and Reference Tracking Strategies for Safety-Critical Robots}
\author{
\IEEEauthorblockN{Pedro Lopes}
\IEEEauthorblockA{INESC TEC\\Porto, Portugal\\pedro.a.lopes@inesctec.pt}
\and
\IEEEauthorblockN{Amit Shivam}
\IEEEauthorblockA{SYSTEC, ISR-Porto, ARISE-LA\\FEUP, Universidade do Porto\\Porto, Portugal\\amitshivam@upwind.pt}
\and
\IEEEauthorblockN{Lu\'is Tiago Paiva}
\IEEEauthorblockA{SYSTEC, ISR-Porto, ARISE-LA\\FEUP, Universidade do Porto\\Porto, Portugal\\ltpaiva@fe.up.pt}}
\begin{document}
\maketitle

\begin{abstract}
Autonomous racing provides a demanding testbed for collision avoidance under tight geometric and computational constraints. Building on a recent conference contribution and a master's dissertation, LACART combines lookahead-based control barrier functions (CBFs), a relaxed control Lyapunov function (CLF), and quadratic programming (QP) to modify nominal vehicle commands for obstacle avoidance and reference tracking. The method handles circular obstacles and track boundaries in a common optimisation layer. Comparative simulation results against Huang \emph{et al.} \cite{r9} show similar full-track paths and lower median computation times: 0.31 versus 0.33 ms per step and 0.023 versus 0.025 ms per QP solve. The dissertation further demonstrates a 1:43-scale vehicle with camera-based estimation and wireless actuation at a reported 25 Hz control rate. We identify the conditions linking lookahead-point clearance to body clearance, illustrate hardware limitations, and propose a recorded demonstration of safety intervention and recovery. The contribution targets ground-robot collision avoidance and real-time safety monitoring.
\end{abstract}

\begin{IEEEkeywords}
Control barrier functions, autonomous vehicles, collision avoidance, runtime assurance, quadratic programming.
\end{IEEEkeywords}

\section{Overview And Contribution}

A nominal path tracker may lead to unsafe motion near obstacles or track limits. A safety filter is often required to ensure safe passage and allow the robot to return to its reference. This abstract synthesises the lookahead CLF--CBF--QP formulation of \cite{r1} and its experimental implementation in the UP2Speed dissertation \cite{r2}. In addition, we compare with recent literature to emphasise the merits of the proposed approach \cite{r9}.

The approach builds on CBF-QP safety filtering \cite{r3} and the runtime-assurance perspective \cite{r4}. Offset-point transformations are established tools for nonholonomic control \cite{r5}, \cite{r6}. CBF filters can also exhibit undesired equilibria \cite{r7}, \cite{r8}, so successful avoidance does not by itself establish freedom from deadlock.
The practical contribution is the integration of steering-sensitive obstacle constraints, track-boundary constraints, and relaxed reference tracking in a small QP, together with a hardware implementation and explicit discussion of its sensing and actuation limits. The comparative simulations demonstrate lower observed computation
times with qualitatively similar trajectories, while the hardware
experiments demonstrate the feasibility of integrating reference
tracking and safety-critical aspects on a miniature vehicle.

\section{Lookahead Safety Filter}
Let the axle position be $p=(x,y)$, heading be $\theta$, and inputs be $u=(v,\omega)$. Define $t=(\cos\theta,\sin\theta)$ and $n=(-\sin\theta,\cos\theta)$. For the unicycle kinematics, the point a distance $l>0$ ahead of the axle satisfies
\begin{equation}\label{eq:point}
p_l=p+lt,\qquad \dot p_l=vt+l\omega n.
\end{equation}
For a static obstacle centred at $o_i$, set $d_i=p_l-o_i$ and $r_{s,i}=r_{\mathrm{obs},i}+r_{\mathrm{rob}}+d_{\min}$. The barrier and its affine input constraint are
\begin{equation}\label{eq:barrier}
\begin{aligned}
h_i&=\|d_i\|^2-r_{s,i}^2,\\
2d_i^T(vt+l\omega n)&\geq-\alpha h_i.
\end{aligned}
\end{equation}
Unlike an axle-centred distance barrier, \eqref{eq:barrier} permits angular velocity to influence safety whenever the lateral component of $d_i$ is nonzero. Track limits are represented by nearby polyline segments using their closest points and zero obstacle radius \cite{r1,r2}. The active segment geometry must be updated as the robot moves.
With positive-definite input weight $W$, relaxation penalty $p_s>0$, and a quadratic tracking-error function $V$, the filter solves
\begin{equation}\label{eq:qp}
\begin{aligned}
\min_{u,\delta\geq0}\quad &\tfrac12(u-u_{\mathrm{nom}})^TW(u-u_{\mathrm{nom}})+p_s\delta^2\\
\text{s.t.}\quad &\eqref{eq:barrier},\quad \dot V\leq-\varepsilon V+\delta,\quad u\in\mathcal U.
\end{aligned}
\end{equation}
The CLF relaxation prioritises barrier constraints when tracking and avoidance conflict.  Lookahead, barrier rate, tracking rate, and relaxation weight tune the balance between early intervention and smooth recovery \cite{r1,r2}.

\section{Comparative Simulation Results}

Reference \cite{r9} uses a time-varying CBF and a CLF-QP for goal-directed unicycle navigation with static and moving obstacles. It also uses an offset between the rear axle and robot centre
to make the barrier constraint depend on both linear and angular
velocity. Accordingly, the distinction here is the track-following integration, boundary handling, and the experimental platform, which establishes the merits of the proposed LACART approach.

\begin{figure}[t]
\centering
\includegraphics[width=\columnwidth]{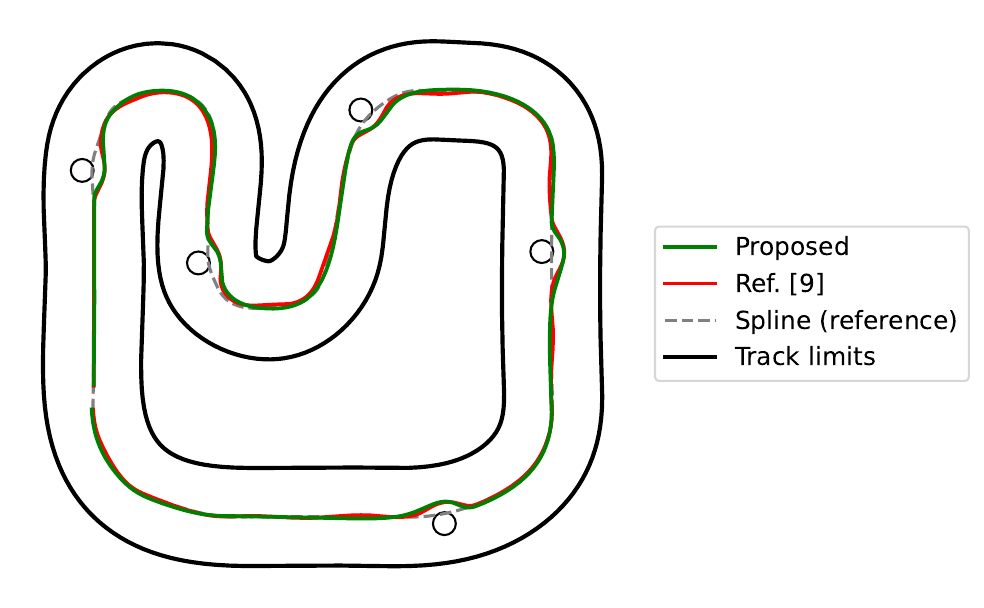}
\caption{Comparative simulation results: proposed (green),  Ref. \cite{r9} (red), spline reference (grey dashed), and track limits (black). Circles denote obstacles. }\label{fig:track}
\end{figure}
\begin{figure}[t]\centering
\includegraphics[width=.8\columnwidth]{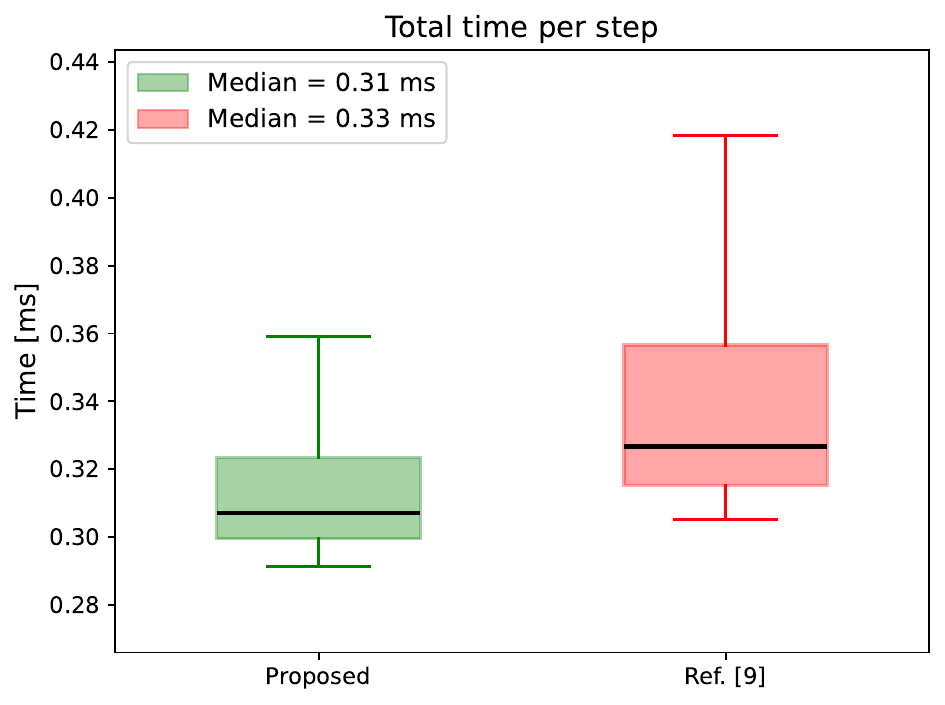}\\
\includegraphics[width=.8\columnwidth]{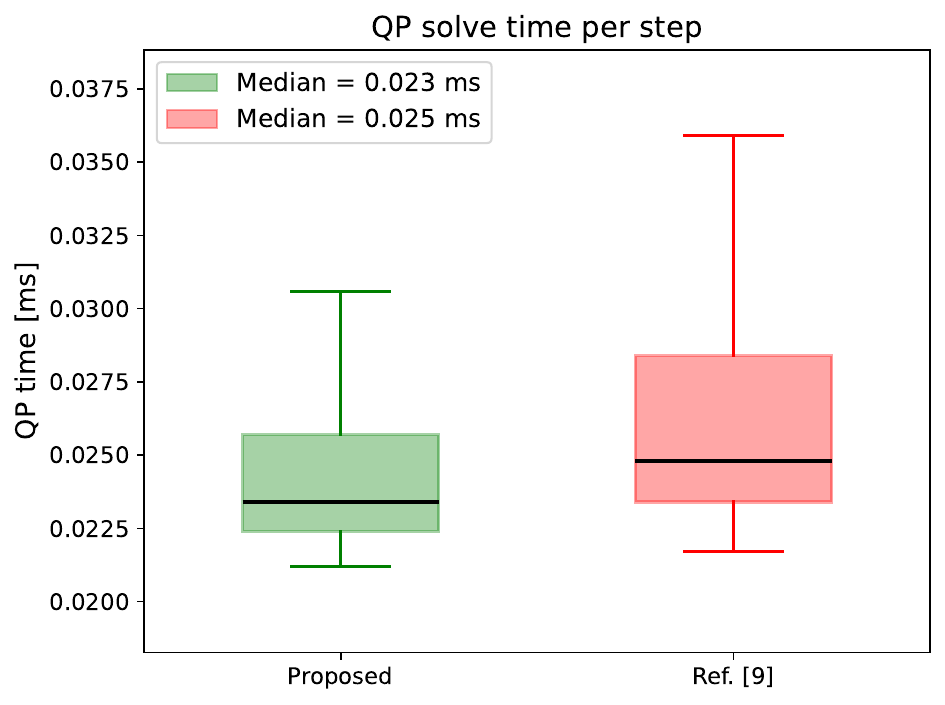}
\caption{Supplied simulation timing distributions: total computation per step (top) and QP solution time (bottom). Median values are the rounded annotations in the original plots; simulation timing is distinct from the hardware timing in Table~\ref{tab:timing}.}\label{fig:times}\end{figure}

Fig.~\ref{fig:track} shows numerical path-tracking trajectories with similar performance. However,
the rounded medians in Fig.~\ref{fig:times} give a total per-step time reduction of (0.33 - 0.31)/0.33 $\approx$ \textbf{6.1\%} and a QP-time reduction of (0.025 - 0.023)/0.025 = \textbf{8.0\%}. These correspond to absolute savings of approximately 20 $\mu$s and 2 $\mu$s, respectively.
The dissertation also investigates offline parameter tuning and relative-motion constraints for multiple vehicles \cite[Chs.~5--6]{r2}. These establish useful extension directions, but are not attributed to the comparison in Figs. 1--2. 
The submission focuses on the lookahead filter and its physical demonstration.

\section{Real-Vehicle Implementation}

The 1:43-scale platform has a rear DC motor, front steering servo, and ESP32-C3 controller. An overhead RunCam Pro 5 camera provides images for OpenCV HSV segmentation. Position is obtained from the marker centroid and heading from successive positions. State estimates reach the external controller through UDP; computed commands reach the car through BLE \cite[Ch.~7]{r2}.
\begin{figure*}[t]
\centering
\begin{tikzpicture}[font=\scriptsize,
  blk/.style={draw,rounded corners=2pt,align=center,minimum height=9.5mm,inner sep=2pt},
  arr/.style={-{Latex[length=1.6mm]},thick},
  lab/.style={font=\scriptsize\itshape,align=center}]
\node[blk,fill=blue!8] (cam) {Overhead camera\\+ HSV segmentation};
\node[blk,fill=blue!8,right=14mm of cam] (est) {State estimate\\$(\hat x,\hat y,\hat\theta)$, latency $\tau$};
\node[blk,fill=yellow!20,right=15mm of est] (nom) {Nominal tracker\\\emph{(untrusted)}};
\node[blk,fill=green!12,right=7mm of nom] (qp) {CLF--CBF--QP filter\\lookahead $l$, margin $\dmin$};
\node[blk,fill=orange!12,right=15mm of qp] (esp) {ESP32-C3\\$\to$ servo, DC motor};
\draw[arr] (cam)--node[above,lab]{T1}(est);
\draw[arr] (est)--node[above,lab]{UDP\\T2}(nom);
\draw[arr] (nom)--(qp);
\draw[arr] (qp)--node[above,lab]{BLE\\T3}(esp);
\node[blk,dashed,fill=red!6,above=6mm of $(nom.north)!0.5!(qp.north)$,minimum width=52mm] (mon) {Runtime monitor \emph{(planned)}: kinematic plausibility, safe stop};
\draw[arr,dashed] (est.north) |- (mon.west);
\draw[arr,dashed] (mon.east) -| (esp.north);
\node[lab,below=1mm of est] {\textbf{T1}  occlusion};
\node[lab,below=1mm of nom,xshift=6mm] {\textbf{T2} replay, delay};
\node[lab,below=1mm of esp] {\textbf{T3} latency};
\end{tikzpicture}
\caption{Architecture of~\cite{r2} with trust boundaries T1--T3.}
\label{fig:arch}
\end{figure*}
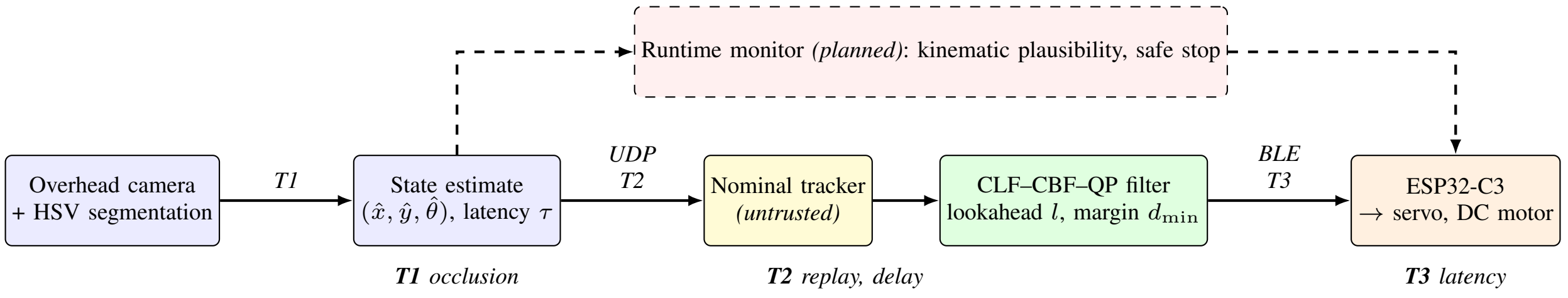
Fig.~\ref{fig:arch} summarises the experimental control architecture, showing the perception, safety-filtering, and actuation stages, the trust boundaries, and the planned runtime monitor. Experiments demonstrate reference following, static-obstacle avoidance, and recovery toward the path (Fig.~\ref{fig:hardware}). The dissertation also reports occasional track-boundary collisions for difficult obstacle placements, attributed to steering limits and delays, and a stop under simultaneous boundary activation. These outcomes document both practical capability and the need to account for the physical vehicle's limits.

\begin{figure}[t]
\centering
\includegraphics[width=.70\columnwidth]{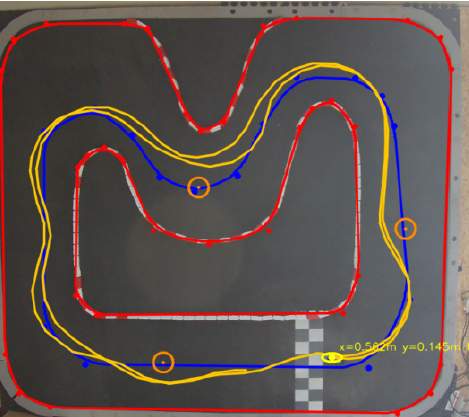}
\caption{Experimental results from \cite[Fig.~7.4, upper panel]{r2}. Blue: reference; yellow: executed path; red: boundaries; orange circles: obstacles.}\label{fig:hardware}
\end{figure}

\begin{table}[t]
\caption{Hardware timing reported in \cite[Table 7.1 and Sec. 7.2.3]{r2}.}\label{tab:timing}
\centering\small
\begin{tabular}{lr}\toprule
Operation / quantity & Reported value\\\midrule
Image acquisition & 6.76 ms\\
Image processing & 27.39 ms\\
State acquisition (UDP) & 28.29 ms\\
QP solution & 7.49 ms\\
BLE communication & 4.34 ms\\
Control-loop period & 40 ms (25 Hz)\\
Mean state-estimation latency & 62.44 ms\\\bottomrule
\end{tabular}
\end{table}

The QP mean is about 19\% of the reported 40 ms loop period. Perception and communication remain substantial costs. Stage averages are not a serial timing allocation: their sum exceeds the reported period. The UP2Speed project GitHub resources are available at
\href{http://github.up2speed.pt}{github.up2speed.pt}.

\section{Safety Conditions and Trust}
For a disk model centred at the axle, safety of the lookahead point must be translated into body safety. If $\|p_l-o\|\geq r_{\mathrm{obs}}+r_{\mathrm{rob}}+d_{\min}$, the triangle inequality gives
\begin{equation}\label{eq:clearance}
\|p-o\|\geq r_{\mathrm{obs}}+r_{\mathrm{rob}}+d_{\min}-l.
\end{equation}
Thus $d_{\min}\geq l$ is a sufficient geometric condition for body clearance for this model. With a bounded lookahead-position error $\eta$, the conservative condition becomes $d_{\min}\geq l+\eta$. This condition relates lookahead-point clearance to body clearance. 
For bounded motion, a state-estimation latency $\tau$ contributes at most $\tau\sqrt{v_{\max}^2+l^2\omega_{\max}^2}$ to lookahead displacement, in addition to estimation error. The measured mean age of 62.44 ms is not a worst-case bound. At 1 m/s, its translational component alone is 6.24 cm.
The real car cannot rotate in place: for a fixed maximum steering
angle, the maximum achievable yaw rate decreases with forward speed. Safety claims consequently depend on feasible actuation,
sufficiently accurate and up-to-date state estimates,
and a feasible barrier problem. Exact head-on geometry can still remove instantaneous angular influence in (2); lookahead does not eliminate every deadlock \cite{r7}, \cite{r8}.

This work addresses robotic safety and real-time intervention. The use of UDP/BLE communication and visual markers motivates
monitoring communication delays and checking the consistency of
state estimates with the vehicle's motion limits. Safe stopping
also requires sufficient clearance to accommodate the experimentally
measured stopping distance.

\section{Demonstration Plan}

A recorded demonstration is proposed on the existing miniature-car platform, with a software-in-the-loop replay as fallback. The display will show the camera view, reference and executed paths, active constraints, commanded inputs, and computation time. Adaptation to the organisers' platform would require interface and actuation validation.

\emph{Scenario 1---Tracking and avoidance:} first follow the obstacle-free reference, then introduce a static obstacle to show safety intervention and recovery. Repeat near a boundary to expose the steering and stopping limits.

\emph{Scenario 2---Lookahead and benchmark:} replay zero- and positive-lookahead simulations, followed by the proposed and \cite{r9} trajectories and timing distributions. This isolates the role of the offset point without implying that \cite{r9} lacks steering control.

\emph{Scenario 3---Planned extension:} introduce controlled frame delays in simulation, then on the car only after validating a conservative stopping envelope. A proposed monitor flags stale or implausible state updates and requests a stop. Report minimum obstacle clearance, tracking error, completion/stopping outcome, and QP time. 

\section{Conclusion}

LACART combines a compact lookahead CLF--CBF--QP with track-boundary handling and a demonstrated miniature-vehicle implementation. Against the supplied implementation of \cite{r9}, the observed median total and QP times decrease by approximately 6.1\% and 8.0\%, with qualitatively similar paths. The principal contribution is the integration of computationally
efficient safety filtering, track-constrained reference tracking,
and experimental validation on a miniature vehicle. By identifying
the effects of body geometry, sensing latency, and actuator limits,
the work provides a concrete basis for evaluating both the
capabilities and limitations of robotic safety intervention. 

\end{document}